\documentclass{article}

\usepackage{microtype}
\usepackage{graphicx}
\usepackage{subfigure}
\usepackage{booktabs} 

\usepackage{hyperref}

\usepackage[accepted]{icml2024}

\usepackage{amsmath}
\usepackage{amssymb}
\usepackage{mathtools}
\usepackage{amsthm}

\usepackage[capitalize,noabbrev]{cleveref}

\theoremstyle{plain}

\theoremstyle{definition}

\theoremstyle{remark}

\usepackage[textsize=tiny]{todonotes}

\usepackage{bm}
\usepackage{subfigure,tikz}
\usepackage{xcolor} 
\usepackage{amsfonts,gensymb}
\DeclareMathOperator*{\argmin}{argmin}
\usepackage{multirow,makecell,tabularx,booktabs,hhline}
\newcolumntype{R}{@{\extracolsep{3cm}}r@{\extracolsep{0pt}}}
\usepackage{nicematrix,xcolor}
\usepackage{colortbl}
\colorlet{Gray}{gray!50}
\usepackage[ruled]{algorithm2e}  
\colorlet{TitleColor}{gray!10}
\usepackage{pifont}

\usepackage{caption}
\makeatletter
\renewcommand{\algocf@makecaption@ruled}[2]{%
  \global\sbox\algocf@capbox{\colorbox{TitleColor}{\hskip\AlCapHSkip
    \parbox[t]{\hsize}{\algocf@captiontext{\strut#1}{\strut#2\strut}}\hskip1.4\algomargin}}
}
\patchcmd{\algocf@makecaption@ruled}{\hsize}{.47\textwidth}{}{} 
\makeatother

\icmltitlerunning{Submission and Formatting Instructions for ICML 2024}
\icmltitlerunning{Exemplar-Free Class Incremental Learning via Incremental Representation}

\begin{document}

\twocolumn[
\icmltitle{Exemplar-Free Class Incremental Learning via Incremental Representation}



\icmlsetsymbol{equal}{*}

\begin{icmlauthorlist}

\icmlauthor{Libo Huang}{ict}
\icmlauthor{Zhulin An}{ict}
\icmlauthor{Yan Zeng}{btbu}
\icmlauthor{Chuanguang Yang}{ict}
\icmlauthor{Xinqiang Yu}{ict}
\icmlauthor{Yongjun Xu}{ict}
\end{icmlauthorlist}

\icmlaffiliation{ict}{Institute of Computing Technology, Chinese Academy of Sciences, Beijing, China.}
\icmlaffiliation{btbu}{School of Mathematics and Statistics, Beijing Technology and Business University, Beijing, China}

\icmlcorrespondingauthor{Zhulin An}{anzhulin@ict.ac.cn}




\icmlkeywords{Machine Learning, ICML}

\vskip 0.3in
]



\printAffiliationsAndNotice{}  

\begin{abstract}
Exemplar-Free Class Incremental Learning (efCIL) aims to continuously incorporate the knowledge from new classes while retaining previously learned information, without storing any old-class exemplars (i.e., samples). For this purpose, various efCIL methods have been proposed over the past few years, generally with elaborately constructed old pseudo-features, increasing the difficulty of model development and interpretation. In contrast, we propose a \textbf{simple Incremental Representation (IR) framework} for efCIL without constructing old pseudo-features. IR utilizes dataset augmentation to cover a suitable feature space and prevents the model from forgetting by using a single L2 space maintenance loss. We discard the transient classifier trained on each one of the sequence tasks and instead replace it with a 1-near-neighbor classifier for inference, ensuring the representation is incrementally updated during CIL. Extensive experiments demonstrate that our proposed IR achieves comparable performance while significantly preventing the model from forgetting on CIFAR100, TinyImageNet, and ImageNetSubset datasets. 
\end{abstract}

\section{Introduction}

\label{sec:intro}
Class Incremental Learning (CIL) is a generalized learning paradigm that enables a model to continually incorporate the knowledge from a sequence of tasks without frogetting~\citep{masana2022class,zhou2023deep}. A dominant way to achieve CIL is \textit{exemplar-based methods} that store exemplars (i.e., samples) from the learned tasks and implicitly simplify CIL into the class-imbalanced learning problem~\cite{rebuffi2017icarl, he2021tale}. Due to its conceptual simplicity and practical effectiveness, CIL has gained a wide range of success in semantic segmentation~\citep{cha2021ssul}, object detection~\citep{shmelkov2017incremental}, and the training of transformers~\citep{wang2022learning}, etc.

One limitation of the exemplar-based CIL methods is their \textit{high memory demand} to retain old samples, which may raise privacy concerns and even violate the government regulations~\citep{masana2022class}. For example, for the widely engaged CIL settings, i.e., CIFAR100 with ResNet32 (and ImageNet100 with ResNet18), storing $20$ exemplars for each learned class would consume $5.85$MB ($287$MB) memory, considerably larger than that of the backbone network, $1.76$MB ($42.6$MB)~\citep{zhou2023deep}.

\begin{figure}[t]
\centering
\includegraphics[width=0.47\textwidth]{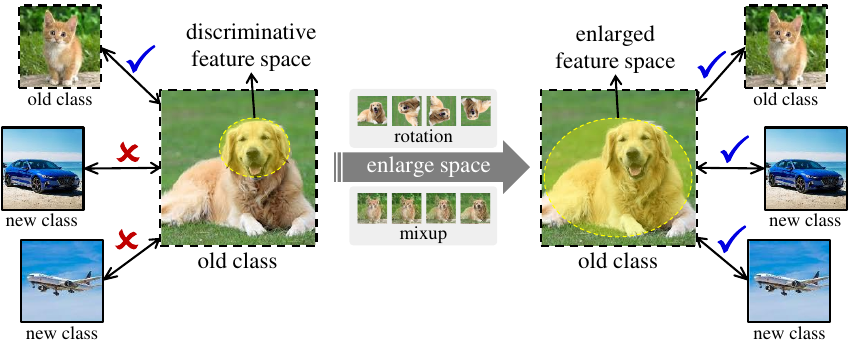}
\caption{Motivation of \textit{Incremental Representation}, that enlarges the feature space with dataset augmentation strategies, e.g., rotation~\citep{gidaris2018unsupervised} or mixup~\citep{zhang2018mixup}. The enlarged feature space enables the feature extractor to incrementally fit the new tasks without any elaborate old feature construction.}
\label{fig:motivation}
\end{figure}

A bunch of \textit{exemplar-free CIL} (efCIL) methods have been proposed to overcome this drawback. \citet{zhu2021prototype} firstly proposed PASS explicitly for efCIL. They split the incremental network into a feature extractor and a classifier, and trained the former with the alignment loss (L2) and the latter with the stored prototypes (feature mean of each old class). Specifically, they employed a L2 loss to prevent the feature extractor from forgetting and augmented each stored prototype with random Gaussian noise to generate sufficient old features, alleviating the biased prediction toward the new classifier head~\citep{zhu2021prototype,zhu2022self}.~\citet{zhu2021class} expanded~\citep{zhu2021prototype, zhu2022self} from the finite old pseudo-features to an infinite number and derived an easy-to-implement upper bound of the loss function to optimize~\citep{wang2021regularizing}. Instead of augmenting the prototype to estimate the old class feature distribution,~\citet{petit2023fetril} assumed feature distributions of all classes are similar and replaced the old prototype with its nearest feature center of new classes to construct the same number of old features. Although we indeed see the constant progress of these works under the exemplar-free condition, neither effective representations nor well-optimized hyper-parameters ensuring their success are easily achievable in practice. Furthermore, the diversity of pseudo-features hinders the emergence of a unified and clear interpretation of the final improvement in efCIL performance.

In this paper, we present a simple yet efficient efCIL framework termed Incremental Representation (IR) and demonstrate that it significantly prevents the model from forgetting with no need for pseudo-features. Our motivation for the proposed IR is illustrated in Fig.\ref{fig:motivation}. We argue that a suitable incremental representation is powerful enough for CIL, which has no need for the elaborately designed strategy of feature generation to balance the biased classifier~\cite{guo2022online}. Based on this argument, which is empirically supported later on, we train a CIL model through space alignment in the preceding layer of the discarded classifier and directly use a 1-near-neighbor (1-NN) classifier for inference. In this way, if we could perfectly construct and align a feature representation that covers a proper feature space, the forgetting along the CIL would considerably be alleviated. That is to say, only the construction and alignment errors account for the CIL model's accuracy, rendering our incremental knowledge acquisition more comprehensible.

For experimental verification, we utilize two data augmentation techniques: data rotation and mixup. Our results show that both data rotation and mixup could construct a feature representation covering a proper feature space, and a single L2 loss for feature alignment already works surprisingly well. Such a simple combined loss saves us from carefully tuning hyper-parameters as previous works do in order to balance the effect of multiple losses. Our main contributions are summarized in three respects\footnote{The source code will be publicly available upon acceptance.}:
\begin{itemize}
    \item We draw a paramount conclusion about the designation of efCIL methods: An initial feature representation covering a proper feature space is essential for efCIL, and we experimentally verify that the dataset rotation or mixup technique is powerful enough to achieve it. 
    \item We develop a general efCIL framework, which empirically turns out to be more straightforward but effective in that it throws away 1) the top incremental classifiers trained on each task and 2) the procedures of elaborately optimizing and constructing old pseudo-features.
    \item We propose a simple and effective efCIL method, IR. IR achieves comparable performance while significantly preventing the model from forgetting, as demonstrated by the extensive experiments on CIFAR100, TinyImageNet, and ImageNetSubset datasets.
\end{itemize}

\section{Related Work} \label{efCIL}
\subsection{Class Incremental Learning}
\textbf{Definition.}\quad
CIL is a broad incremental learning paradigm compared with the remaining two, task-incremental learning and domain-incremental learning~\citep{van2022three, hsu2018re, van2018three}. Their crucial difference lies in the inference period. When presented with a test sample, CIL must infer the task identity and the corresponding class number. In contrast, domain-incremental learning only infers the class number, while task-incremental learning requires prior knowledge of the sample's task identity for inference. The primary challenge of CIL is catastrophic forgetting~\citep{goodfellow2013empirical, mccloskey1989catastrophic}, which refers to a significant decline in performance on previously learned tasks when the model is trained on a new task.

\noindent\textbf{Methods.}\quad
Existing CIL methods mainly fall into three categories, parameter isolation, regularization-based, and rehearsal-based methods~\citep{huang2022lifelong, masana2022class}. Parameter isolation methods maintain either a set of parameters or a sub-network for each learned task~\citep{rusu2016progressive,yan2021dynamically}. Regularization-based methods prevent the crucial knowledge of learned tasks from changing by either estimating parameters' importance~\citep{kirkpatrick2017overcoming,aljundi2018memory} or projecting gradients to a subspace~\citep{zeng2019continual,wang2021training}. Rehearsal-based methods reply the pseudo samples from either incremental generative networks~\citep{shin2017continual,wu2018memory} or incremental discriminative networks~\citep{yin2020dreaming, smith2021always}; or directly replay the stored old exemplars~\citep{rebuffi2017icarl,hou2019learning}. 

Although numerous CIL strategies exist, methods with exemplars generally perform better than those without exemplars~\cite{liu2020mnemonics,zhao2020maintaining,yan2021dynamically,liu2021rmm}. Unlike the dominant exemplar-based CIL methods in the rehearsal-based category, we study a more challenging CIL problem under the exemplar-free scenario in this paper, where data privacy is strictly demanded, and sample storing is rigorously prohibited.

\begin{figure*}[t]
\centering
\includegraphics[width=0.98\textwidth]{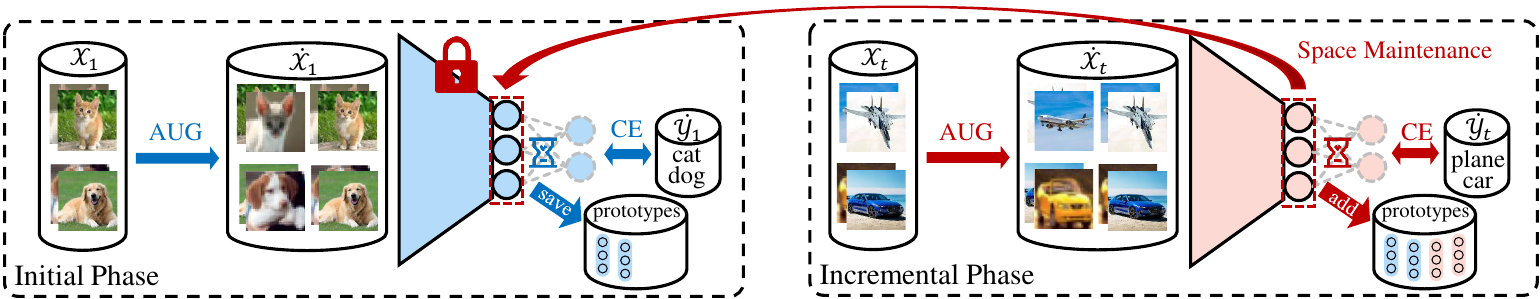}
\caption{Framework of \textit{Incremental Representation}, where AUG and CE are intra-phrase processes while space maintenance is an inter-phases process. AUG enlarges the feature space by augmenting the input dataset with different strategies (e.g., rotation~\citep{gidaris2018unsupervised} or mixup~\citep{zhang2018mixup}), and the transient cross-entropy (CE) plastically enables the intra-phase features discriminative. Space maintenance stably transfers the previous knowledge from the frozen feature extractor to the current trainable one, and negligible storage is used to store the prototypes for the 1-NN classifier.}
\label{fig:method}
\end{figure*}

\subsection{Exemplar-Free Class Incremental Learning} \label{efCIL}
\noindent\textbf{Taxonomy.}\quad
Compared with CIL, there has been relatively less research on efCIL. On the one hand, some typical regularization- and rehearsal-based CIL methods belong to efCIL, such as LwF~\citep{li2016learning}, EWC~\citep{kirkpatrick2017overcoming}, LwM~\citep{dhar2019learning}, etc. On the other hand, we organize the recent efCIL methods taxonomically in two respects, compensation-centric and prototype-centric. 

\noindent\textbf{Methods.}\quad
Compensation-centric efCIL methods estimate and compensate the semantic drift arising in the incremental phases based on the current-task's samples~\citep{yu2020semantic,toldo2022bring}. Prototype-centric efCIL methods design various old-feature generative strategies with the stored old class means (i.e., prototype). Note that the compensation-centric efCIL methods commonly perform worse than prototype-centric ones~\citep{petit2023fetril,zhu2022self}. Thus we mainly pay attention to these prototype-centric efCIL methods in this paper. Concretely, \citet{zhu2021prototype} augmented each stored prototype with random Gaussian noise to generate sufficient old features, alleviating the biased prediction toward the new classifier head~\citep{zhu2021prototype}. Based on \citet{zhu2021prototype}, \citet{zhu2022self} introduced a prototype attention mechanism to assign different prototypes with different weights. \citet{zhu2021class} expanded~\citep{zhu2021prototype, zhu2022self} from the finite old pseudo-features to an infinite number and derived an easy-to-implement upper bound of the loss function to optimize~\citep{wang2021regularizing}. Instead of augmenting the prototype to estimate the old-classes' feature distribution,~\citet{petit2023fetril} assumed feature distributions of all classes are similar and used the prototype in place of its nearest feature center of new classes to construct old features.

Similar to prototype-centric efCIL methods, we also store one prototype for each learned class. However, we are ultimately different from the existing efCIL methods in (1) we are concentrated on designing an incremental representation framework from the perspective of enlarging discriminative feature space rather than semantic drift estimation or compensation, and (2) we engage the prototypes for the nonparametric 1-NN classifier with no old-feature generative strategies as the prototype-centric efCIL methods.

\section{Method} \label{sec:met}
\noindent\textbf{Problem Statement.}\quad
The objective of efCIL is to learn a sequence of tasks without storing old samples during incremental training and without knowing the task identity of the test samples during inference. We refer to each learning task in the sequence as a \textit{phase}. There are $(T+1)$ learning phases characterized by datasets $\mathbb{D}=\{\mathbb{D}_t\}_{t=0}^T$, where $\mathbb{D}_t$ represents the labeled dataset accessible to the model at phase $t$. It contains samples $\left(\bm{x}_{t,i}, y_{t,i}\right)$ with $i$ ranging from $1$ to $N_T$, and the corresponding class labels $y_{t,i}\in \mathbb{C}_t$. $N_t$ and $\mathbb{C}_t$ represent the number of samples and the set of classes at phase $t$, respectively. Importantly, the classes in different phases are disjoint, i.e., $\mathbb{C}_i\cap\mathbb{C}_j=\emptyset$ for all $i\neq j$. In the incremental phase $t>0$, the efCIL model aims to learn the new task $t$ plastically while stably preserving the knowledge acquired from the previous phases.

\noindent\textbf{Framework Overview.}\quad
Our proposed incremental representation framework, depicted in Fig.\ref{fig:method}, involves two types of phases. In the initial learning phase, IR augments the dataset with different strategies (AUG), e.g., rotation~\citep{gidaris2018unsupervised} or mixup~\citep{zhang2018mixup}, to enlarge the feature space, and engages the transient cross-entropy (CE) loss to make the intra-phase features discriminative. After training, we discard the top classifier trained with CE and instead store a prototype of each class in a negligible buffer for the final nonparametric 1-NN classifier. In the incremental learning phase, we additionally introduce a space maintenance strategy that transfers previous knowledge from the frozen feature extractor to the current trainable one. We update the prototypes of current classes in the buffer during this phase.

\subsection{Proposed Incremental Representation} \label{subsec:proposed}
IR framework enlarges the representation feature space by using the dataset augmentation techniques:\textit{rotation}~\citep{gidaris2018unsupervised} or \textit{mixup}~\citep{zhang2018mixup}:

$\bullet$\quad
\textit{Rotation} augments ${\mathbb{D}_t=\left\{\left(\bm{x}_{t,i}, y_{t,i}\right)|_{i=1}^{N_t},\ y_{t,j}\in \mathbb{C}_t\right\}}$ to ${\dot{\mathbb{D}}_t =\left\{(\dot{\bm{x}}_{t,i},\dot{y}_{t,i})|_{i=1}^{4 N_t},\ \dot{y}_{t,j} \in \dot{\mathbb{C}}_t\right\}}$ by rotating each sample with multiples of $90$ degrees (i.e., $0\degree $, $90\degree $, $180\degree $, or $270\degree $), where $|\dot{\mathbb{C}}_t|=4\times|\mathbb{C}_t|$, and $|\cdot|$ is the cardinal operator~\citep{gidaris2018unsupervised}.

$\bullet$\quad
\textit{Mixup} augments $\mathbb{D}_t$ to $4 N_t$ samples. It randomly interpolates 2 different-class samples, $\bm{x}_{t,i}$ and $\bm{x}_{t,j}$ ($i\neq j, y_{t,i}\neq y_{t,j}$), with an interpolation factor $\alpha\in[0.4, 0.6]$ to create a new sample $\dot{\bm{x}}_{t, k}=\alpha \bm{x}_{t,i} +(1-\alpha) \bm{x}_{t,j}$, and assigns the interpolated sample to a new category $\dot{y}_{t,k}\neq y_{t,i}\neq y_{t,j}$, with
$|\dot{\mathbb{C}}_t|=\left(|\mathbb{C}_t|+|\mathbb{C}_t|^{2}\right)/2$~\citep{zhang2018mixup}.

Similar to~\citet{rebuffi2017icarl,hou2019learning,zhao2020maintaining,zhu2021prototype,toldo2022bring}, we split the network in the $t$-th phase into two separates, feature extractor $\mathcal{F}_{\bm{\theta}_t}$ and classifier $\mathcal{H}_{\bm{\psi}_t}$ parameterized by $\bm{\theta}_t$ and $\bm{\psi}_t$, respectively.

\noindent\textbf{Plastic IR for New Phase.}\quad
To enable the efCIL model to learn classes in the new phase plastically, we train the model with the following cross-entropy loss on the dataset $\dot{\mathbb{D}}_t$,
\begin{align*} 
    \mathcal{L}_{ce}(\bm{\theta}_t, \bm{\psi}_t) =-\frac{1}{4N_t}\sum_{i=1}^{4N_t}\sum_{j=1}^{|\dot{\mathbb{C}}_t|}\delta_{\dot{y}_{t,i} =j}\log\bm{p}_{t,j}(\dot{\bm{x}}_{t,i};\tau),
\end{align*}
where $\delta_{\dot{y}_{t,i}=j}$ is the indicator function. $\bm{p}_{t,j}(\dot{\bm{x}}_{t,i}; \tau)$ is the softmax ($\sigma$) value of logits on the $j$-th class head, 
\begin{align*}
    \bm{p}_{t,j}(\dot{\bm{x}}_{t,i}; \tau ) 
    & =\sigma \left(\mathcal{H}_{\bm{\psi}_t}\left(\mathcal{F}_{\bm{\theta}_t}(\dot{\bm{x}}_{t,i})\right);\tau\right) \\
    & =\frac{e^{\left.\left(\mathcal{H}_{\bm{\psi}_t}\left(\mathcal{F}_{\bm{\theta}_t}(\dot{\bm{x}}_{t,i})\right)\right)_j\right/\tau}}{\sum_{k=1}^{|\dot{\mathbb{C}}_t|}e^{\left.\left(\mathcal{H}_{\bm{\psi}_t}\left(\mathcal{F}_{\bm{\theta}_t}(\dot{\bm{x}}_{t,i})\right)\right)_k\right/\tau }},
\end{align*}
where $\tau>0$ is the temperature hyper-parameter, and it controls the smoothness of output distribution. In general, higher temperatures tend to produce flatter outputs, introducing a more robust but less precise estimation of the category distribution~\citep{hinton2015distilling}. 

\textit{\textbf{Discussion.}} The cross-entropy loss, $\mathcal{L}_{ce}(\cdot)$, 
is commonly used in deep learning~\citep{murphy2012machine}, and the associated temperature parameter has been discussed in the knowledge distillation and contrastive learning communities~\citep{hinton2015distilling, chen2020simple}. However, fewer studies have been conducted on how temperature affects the training of deep models. 
A higher temperature tends to produce a more flattened semantic representation, leading to a poorer performance of the top classifier. IR's final prediction is determined by a 1-NN, which is independent of the top classifier. Consequently, tuning the temperature parameter if crucial for optimizing IR outcomes.

\begin{table*}[!t]
    \caption{Average incremental accuracy (IA$\uparrow$) along with its {\tiny$\pm std$} of several CIL methods on CIFAR100, TinyImageNet, ImageNetSubset. We emphasize our reported \textbf{best} results in bold and the \underline{second best} underlined. 
    $E$ and $P$ indicate the number of exemplars and incremental phases, respectively. }  
    \label{tab:ia}
    \centering
    \resizebox{.998\linewidth}{!}{
    \begin{NiceTabular}{ll lll lll l}[code-before=\rowcolors{}{}{Gray} \rowcolor{TitleColor}{1,2}]
    \toprule
        \multicolumn{2}{c}{\multirow{2}*{CIL Method}}  &  \multicolumn{3}{c}{CIFAR100}  &  \multicolumn{3}{c}{TinyImageNet}  &  ImageNetSubset  \\
        \cmidrule(lr){3-5}  \cmidrule(lr){6-8}  \cmidrule(lr){9-9}
            &    &  $P=5$  &  $P=10$  &  $P=20$  &  $P=5$  &  $P=10$  &  $P=20$  &  $P=10$  \\
        \midrule
        \multicolumn{2}{c}{Lower Bound}  &  23.07{\tiny$\pm0.12$}  &  12.91{\tiny$\pm0.08$}  &  7.94{\tiny$\pm0.02$}  &  19.26{\tiny$\pm0.29$}  &  10.99{\tiny$\pm0.18$}  &  5.92{\tiny$\pm0.11$}  &  13.31{\tiny$\pm0.12$}  \\
        \multicolumn{2}{c}{Upper Bound}  &  75.81{\tiny$\pm0.45$}  &  75.77{\tiny$\pm0.28$}  &  76.11{\tiny$\pm0.39$}  &  60.39{\tiny$\pm0.69$}  &  60.17{\tiny$\pm0.77$}  &  59.93{\tiny$\pm0.82$}  &  79.08{\tiny$\pm2.73$}  \\
        \cmidrule(lr){1-9}
        \multirow{4}{*}{ \rotatebox[origin=c]{90}{\makecell{($E=2k$)}} }
          &  iCaRL-CNN  &  54.96{\tiny$\pm0.61$}  &  51.32{\tiny$\pm0.66$}  &  48.01{\tiny$\pm0.33$}  &  40.31{\tiny$\pm0.75$}  &  36.40{\tiny$\pm0.46$}  &  33.31{\tiny$\pm0.61$}  &  51.04{\tiny$\pm0.7$}  \\
          &  iCaRL-NME  &  62.29{\tiny$\pm0.54$}  &  57.46{\tiny$\pm0.45$}  &  53.74{\tiny$\pm0.67$}  &  49.34{\tiny$\pm0.85$}  &  44.60{\tiny$\pm0.79$}  &  40.05{\tiny$\pm0.70$}  &  57.53{\tiny$\pm0.48$}  \\
          &  WA  &  60.33{\tiny$\pm0.31$}  &  53.78{\tiny$\pm0.21$}  &  47.7{\tiny$\pm0.13$}  &  49.01{\tiny$\pm0.71$}  &  42.64{\tiny$\pm0.55$}  &  33.02{\tiny$\pm0.55$}  &  57.02{\tiny$\pm0.7$}  \\
          &  DER  &  67.78{\tiny$\pm0.68$}  &  64.95{\tiny$\pm0.91$}  &  61.3{\tiny$\pm1.37$}  &  54.77{\tiny$\pm0.39$}  &  53.21{\tiny$\pm0.12$}  &  51.77{\tiny$\pm0.24$}  &  66.54{\tiny$\pm0.42$}  \\
        \cmidrule(lr){1-9}
        \multirow{8}{*}{ \rotatebox[origin=c]{90}{\makecell{efCIL ($E=0$)}} }
          &  EWC  &  24.65{\tiny$\pm0.12$}  &  13.66{\tiny$\pm0.11$}  &  8.39{\tiny$\pm0.14$}  &  27.29{\tiny$\pm0.89$}  &  17.15{\tiny$\pm1.69$}  &  8.20{\tiny$\pm0.70$}  &  43.72{\tiny$\pm1.03$}  \\
          &  LwF.MC  &  58.49{\tiny$\pm1.42$}  &  34.98{\tiny$\pm1.49$}  &  23.53{\tiny$\pm1.82$}  &  47.29{\tiny$\pm0.50$}  &  36.12{\tiny$\pm0.38 $}  &  17.17{\tiny$\pm0.92$}  &  42.25{\tiny$\pm1.95 $}  \\
          &  PASS  &  66.09{\tiny$\pm0.16$}  &  61.61{\tiny$\pm0.84$}  &  58.4{\tiny$\pm0.94$}  &  51.03{\tiny$\pm0.52$}  &  48.29{\tiny$\pm0.21$}  &  43.59{\tiny$\pm1.37$}  &  67.39{\tiny$\pm0.72$}  \\  
          &  IL2A  &  66.60{\tiny$\pm0.19$}  &  60.51{\tiny$\pm0.61$}  &  59.72{\tiny$\pm0.94$}  &  51.42{\tiny$\pm1.02$}  &  45.20{\tiny$\pm0.72$}  &  35.55{\tiny$\pm0.73$}  &  64.03{\tiny$\pm0.99$}  \\  
          &  SSRE  &  65.88{\tiny$\pm0.42$}  &  65.04{\tiny$\pm0.55$}  &  61.70{\tiny$\pm0.72$}  &  50.39{\tiny$\pm0.81$} &  48.93{\tiny$\pm0.71$}  &  48.17{\tiny$\pm0.66$}  &  67.69{\tiny$\pm0.75$} \\
          &  FetrIL  &  65.34{\tiny$\pm0.17$}  &  64.44{\tiny$\pm0.18$}  &  $\underline{61.85}${\tiny$\pm0.81$}  &  $51.39${\tiny$\pm1.13$}  &  $50.77${\tiny$\pm1.05$}  &  $\bm{48.11}${\tiny$\pm1.14$}  &  $\underline{69.29}${\tiny$\pm0.71$}  \\  
          \cmidrule(lr){2-9}
          &  IR$_r$  &  $\underline{67.11}${\tiny$\pm1.05$}  &  $\bm{65.88}${\tiny$\pm1.23$}  &  $\bm{62.01}${\tiny$\pm1.25$}  &  $\bm{52.64}${\tiny$\pm0.40$}  &  $\bm{52.55}${\tiny$\pm0.41$}  &  $\underline{47.88}${\tiny$\pm0.16$}  &  $\bm{69.32}${\tiny$\pm1.17$}  \\  
          &  IR$_m$  &  $\bm{67.22}${\tiny$\pm0.83$}  &  $\underline{65.68}${\tiny$\pm0.71$}  &  61.05{\tiny$\pm0.84$}  &  \underline{52.11}{\tiny$\pm0.6$}  &  \underline{50.83}{\tiny$\pm0.33$}  &  46.63{\tiny$\pm0.72$}  &  68.03{\tiny$\pm0.57$}  \\  
        \bottomrule
    \end{NiceTabular}
    }
\end{table*}

\begin{algorithm}[!t]  
\caption{IR training algorithm}\label{alg:ir}
\DontPrintSemicolon
\KwIn{Random initialized model $\left\{\mathcal{H}_{\bm{\psi}_0}, \mathcal{F}_{\bm{\theta}_0}\right\}$; $(T+1)$ learning phases data $\{\mathbb{D}_t\}_{t=0}^T$;}
\KwOut{Model $\left\{\mathcal{H}_{\bm{\psi}_T}\right\}$; prototypes $\{S_T\}$;}
\SetKwBlock{Begin}{function}{end function}
\Begin($ \text{IR}{(} \mathcal{H}_{\bm{\psi}_0}, \mathcal{F}_{\bm{\theta}_0}, \{\mathbb{D}_t\}_{t=0}^T {)}$)
{   
    Augment $\mathbb{D}_0$ to $\mathbb{\dot{D}}_0$ via \textit{Rotation} (or \textit{Mixup}); \;
    Train $\left\{\mathcal{H}_{\bm{\psi}_0}, \mathcal{F}_{\bm{\theta}_0}\right\}$ with Eq.\eqref{eq:obj0}; \;
    Calculate and save $\{S_0\}$ with Eq.\eqref{eq:pro}; \;
    \For{$t = 1,...,T$}
    {
        Randomly initialize $\bm{\theta}_t$, and $\bm{\psi}_t \leftarrow \bm{\psi}_{t-1}$;  \;
        Augment $\mathbb{D}_t$ to $\mathbb{\dot{D}}_t$ via \textit{Rotation} (or \textit{Mixup}); \;
        Train $\left\{\mathcal{H}_{\bm{\psi}_t}, \mathcal{F}_{\bm{\theta}_t}\right\}$ with Eq.\eqref{eq:obj1}; \;
        Calculate prototypes $\bm{s}$ on $\mathbb{D}_t$ with $\mathcal{H}_{\bm{\psi}_t}$ and Eq.\eqref{eq:pro} accordingly; \;
        Update $\{S_t\} \leftarrow \{S_{t-1}\} \cup \{\bm{s}\}$; \;
    }
  \Return{$\mathcal{H}_{\bm{\psi}_T} \ \mathrm{and}\ S_T$}
}
\end{algorithm}

\noindent\textbf{Stable IR for Old Phases.}\quad
To make the efCIL model stably maintain the old knowledge acquired from previous tasks, we regularize the output features from the current trainable network by those from the frozen feature extractor, with the following space maintenance loss on $\dot{\mathbb{D}}_t$,
\begin{align} \label{eq:sm}
    \mathcal{L}_{sm}(\bm{\theta}_t) =\frac{1}{4N_t}\sum_{i=1}^{4 N_t}\left\Vert \mathcal{F}_{\bm{\theta}_t}(\dot{\bm{x}}_{t,i})-\mathcal{F}_{\bm{\theta}_{t-1}}(\dot{\bm{x}}_{t,i})\right\Vert_p,
\end{align}
where $\Vert\cdot\Vert_p$ is the $L_p$-norm regularization.

\textit{\textbf{Discussion.}} Both compensation-centric and prototype-centric efCIL methods detailed in Section~\ref{efCIL} align the outputs from previous models with those from current models implicitly or explicitly~\citep{yu2020semantic,zhu2021class,zhu2022self}. It motivates us that the above space maintenance loss may account for the model's anti-forgetting character, while there is no need for semantic drift compensation post-processing operations or elaborate prototype-based generation of old features. Because (1) this regularization term is relatively strict between two kinds of outputs~\citep{boyd2004convex}, and (2) $\mathcal{L}_{sm}$ assumes that performing space maintenance on the current data $\dot{\mathbb{D}}_t$ can restore the feature space of the previous data $\{\dot{\mathbb{D}}_i\}_{i=1}^{t-1}$ built from old feature extractor, we argue that enlarging the feature space along with the above space maintenance is essential for IR.

\noindent\textbf{Integrated Objective.}\quad
We outline the whole integrated objective of IR in Algorithm~\ref{alg:ir}. In the initial phase $0$, the model has no acquired knowledge. We train the network on the augmented dataset $\mathbb{\dot{D}}_0$ with the following objective, 
\begin{align} \label{eq:obj0}
    \argmin_{\bm{\theta}_0, \bm{\psi}_0}\mathcal{L}_{ce}(\bm{\theta}_0, \bm{\psi}_0).
\end{align}
Until the model is well-trained, we can obtain each class's prototype $S_{0,i}$, $i\in\mathbb{C}_0$ in the dataset, $\mathbb{D}_0$, by calculating the mean of the intra-class features,
\begin{align} \label{eq:pro}
    S_{0,i} = \frac{\sum_{j=1}^{N_{0}} \delta_{y_{0,j} =i} \mathcal{F}_{\bm{\theta}_0}(\bm{x}_{0,j})}{\sum_{j=1}^{N_{0}} \delta_{y_{0,j} =i}}.
\end{align}
In the incremental phases $t>0$, we train the the incremental model $\left\{\mathcal{H}_{\bm{\psi}_t}, \mathcal{F}_{\bm{\theta}_t}\right\}$ on $\mathbb{\dot{D}}_t$ with the objective,
\begin{align} \label{eq:obj1}
    \argmin_{\bm{\theta}_t, \bm{\psi}_t} \left\{\mathcal{L}_{ce}(\bm{\theta}_t, \bm{\psi}_t) + \lambda \mathcal{L}_{sm}(\bm{\theta}_t)\right\},
\end{align}
where the parameter $\lambda>0$ controls the trade-off between the plasticity ($\mathcal{L}_{ce}$) and the stability ($\mathcal{L}_{sm}$). We then update $\{S_t\}$ by unifying $\{S_{t-1}\}$ with the new class's prototype calculated on $\mathbb{D}_t$ with $\mathcal{F}_{\bm{\theta}_t}$ and Eq.\eqref{eq:pro} accordingly. Finally, given a test sample $x\in\{\mathbb{D}_i\}_{i=0}^t$ from the tasks seen so far, we assign its label with the most similar prototype by the 1-NN nonparametric classifier~\citep{bishop2006pattern}.

\begin{figure*}[!t]
\centering
\includegraphics[width=0.98\linewidth]{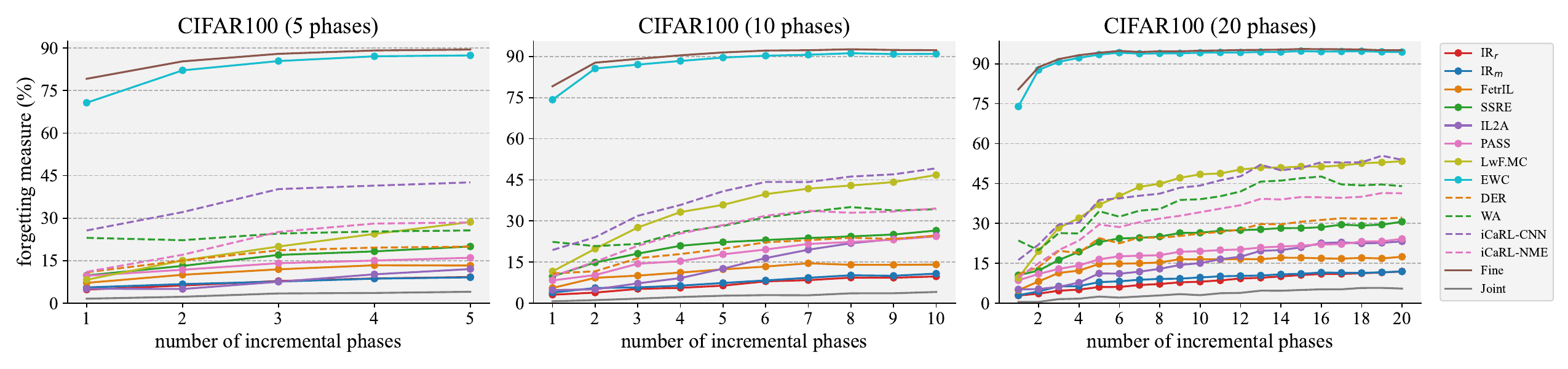}
\caption{Forgetting measures on CIFAR100 for different phrases. Solid lines present efCIL methods, and dashed lines present data replay-based methods.}
\label{fig:fgt_cifar}
\end{figure*}
\begin{figure*}[!t]
\centering
\includegraphics[width=0.98\linewidth]{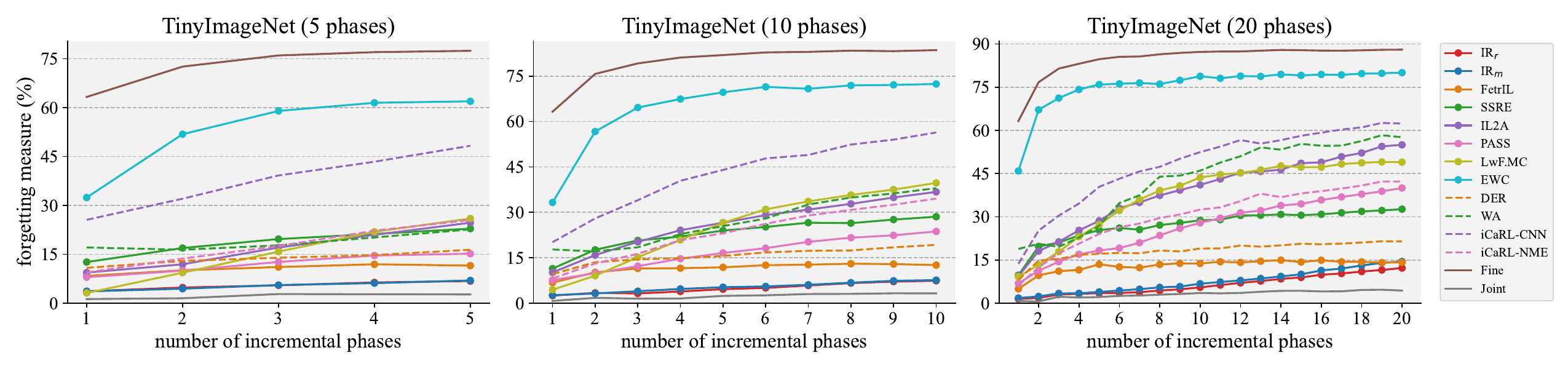}
\caption{Forgetting measures on TinyImageNet for different phrases.}
\label{fig:fgt_tiny}
\end{figure*}
\begin{figure}[!t]
\centering
\includegraphics[width=0.98\linewidth]{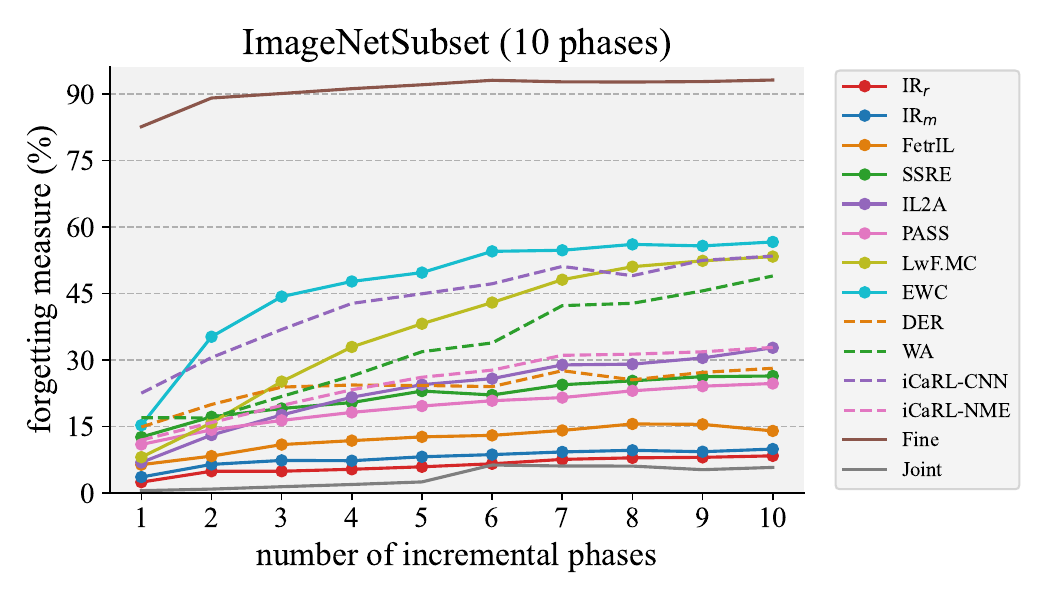}
\caption{Forgetting measures on ImageNetSubset for the setting of $10$ incremental phases.}
\label{fig:fgt_imagenet100_10}
\end{figure}

\section{Experiment}
\subsection{Evaluation Protocol}

\noindent\textbf{Datasets.}\quad
We conduct experiments on CIFAR100, TinyImagenet, and ImageNetSubset datasets~\citep{masana2022class,russakovsky2015imagenet,krizhevsky2009learning}. CIAFR100 contains $50$k training images with $500$ instances per class, and $10$k testing images with $100$ instances per class. Each image has $32\times 32$ pixels. We first pad each image with $4$ pixels, then randomly/center crop to $32\times32$ for training/testing. TinyImageNet contains a total of $200$ classes with $500$ training images and $50$ testing images for each class. Each image has $64\times64$ pixels. We first pad each image with $4$ pixels, then randomly/center crop to $64\times64$ for training/testing. ImageNetSubset comprises $100$ randomly chosen classes from ImageNet~\cite{russakovsky2015imagenet}. 
We resize each image to $256\times256$ pixels before randomly/center cropping to $224\times224$ for training/testing.

\noindent\textbf{Settings.}\quad
Following~\citep{zhou2023deep, zhu2022self}, we engage ResNet-18~\citep{he2016deep} as our backbone network, and train it from scratch for all experiments. We use the Adam optimizer~\citep{kingma2014adam} with $100$ training epochs and set the initial learning rate to $0.001$, which is reduced by multiplying $0.1$ at $45$ and $90$ epochs. The batch size is $64$. We train the model on half of the classes for the initial phase and equal classes in the remaining incremental phases~\cite{zhu2021class}. For example, $P=5$ of CIFAR100 means that the dataset is split into one initial phase with $50$ classes and $5$ incremental phases, each containing $10$ classes. We repeat each experiment with 3 different random seeds and report the final average results. 

\noindent\textbf{Evaluation Metrics.}\quad
We employ \textit{average incremental accuracy} and \textit{forgetting measures}~\citep{rebuffi2017icarl,chaudhry2018riemannian,hou2019learning,yu2020semantic} as evaluation metrics. High average incremental accuracy (IA$\uparrow$), or low forgetting measure (F$\downarrow$) demonstrate a good CIL performance. IA is a single number indicator defined as IA$=\frac{1}{T}\sum_{i=1}^T{A_i}$, where $A_i=\frac{1}{i}\sum_{j=1}^i{a_{i,j}}$, and $a_{i,j}\in[0, 1]$ is the accuracy at the $j$-th task of the model that has been incrementally trained from task $1$ to task $i$, and $i\geq j$. IA is an absolute metric highly related to the result achieved in the initial phase~\citep{petit2023fetril, hayes2020lifelong}. F at the $i$-th task is defined as F$_i=\frac{1}{i-1}\sum_{j=1}^{i-1}f_{i,j}$, where $f_{i,j}=\max_{k\in \{1,...,i-1\}}(a_{k,j}-a_{i,j})\in[-1, 1]$, $\forall i>j$, indicating the maximum difference between the accuracy achieved in the learned phases and the accuracy in the current phase. It is a relative metric that traces all learning phases of every single task~\citep{chaudhry2018riemannian}.


\textbf{Comparison Methods.}\quad
Our proposed efCIL method with rotation (IR$_r$) or mixup (IR$_m$), does not store any old exemplars for rehearsal when learning new classes. Thus, we mainly compare IR with several efCIL methods: EWC~\citep{kirkpatrick2017overcoming}, LwF.MC~\citep{rebuffi2017icarl}, PASS~\citep{zhu2021prototype}, IL2A~\citep{zhu2021class}, SSRE~\citep{zhu2022self}, and FetrIL~\citep{petit2023fetril}. In addition, we compare IR with three state-of-the-art exemplar-based CIL methods: iCaRL~\citep{rebuffi2017icarl}, WA~\citep{zhao2020maintaining}, and DER~\citep{yan2021dynamically}. We follow~\citet{rebuffi2017icarl, yan2021dynamically} to select the stored $2000$ exemplars (abbr. $E=2k$) for the learned class by using the `herding' technique. For iCaRL, its CNN predictions and nearest-mean-of-exemplars classification are denoted as iCaRL-CNN and iCaRL-NME, respectively. 

\begin{table*}[!t]
  \caption{Ablation experiments under $\bm{5}$ incremental phases on CIFAR100. We report the average accuracy of the initial phase (ini.$0$) and the $10$ incremental phases (inc.$x$), the average incremental accuracy (IA$\uparrow$), and the average forgetting measure (Avg.F$\downarrow$). If not marked, the IR framework has not engaged the transient cross-entropy (Tra.CE), space maintenance (Spa.M), or dataset augmentation (AUG).}
  \label{tab:abl_5}
  \centering
    \begin{NiceTabular}{c | ccc | llllll | ll}[code-before = \rowcolors{}{}{Gray} \rowcolor{TitleColor}{1}]
    \toprule
    Method & Tra.CE & Spa.M & AUG & ini.$0$ & inc.$1$ & inc.$2$ & inc.$3$ & inc.$4$ & inc.$5$ & IA$\uparrow$ & Avg.F$\downarrow$ \\
    \midrule
    Variant 1  & \checkmark & - & - &  79.09  &  18.26  &  15.35  &  13.58  &  12.07  &  10.95  &  24.88  &  84.94  \\
    Variant 2  & - & \checkmark & - &  79.09  & 71.31 & 65.35 & 60.83 & 57.41 & 54.12 & 64.68 & 7.542 \\
    Variant 3  & \checkmark & - & \checkmark &  79.29  &  20.02 & 15.86 & 13.87 & 12.40 & 11.34 & 25.46 & 82.65 \\
    Variant 4  & - & \checkmark & \checkmark &  79.29  & 70.67 & 65.80 & 62.10 & 58.83 & 56.01 & 65.45 & 7.38 \\
    \midrule
    IR$_r$ & \checkmark & \checkmark & \checkmark & 79.29 & 72.65 & 67.81 & 64.08 & 60.78 & 58.03 & 67.11 & 7.36 \\
    \bottomrule
    \end{NiceTabular}
\end{table*}
\begin{table*}[!t]
  \caption{Ablation experiments under $\bm{10}$ incremental phases on CIFAR100. Other notations are same as that of Tab.\ref{tab:abl_5}.}
  \label{tab:abl_10}
  \centering
  \resizebox{.998\linewidth}{!}{
    \begin{NiceTabular}{c | lllllllllll | ll}[code-before = \rowcolors{}{}{Gray} \rowcolor{TitleColor}{1}]
    \toprule
    Method & ini.$0$ & inc.$1$ & inc.$2$ & inc.$3$ & inc.$4$ & inc.$5$ & inc.$6$ & inc.$7$ & inc.$8$ & inc.$9$ & inc.$10$ &IA$\uparrow$ & Avg.F$\downarrow$ \\
    \midrule
    Variant 1 & 79.09 &  11.93 & 10.44 & 9.59 & 8.84 & 8.52 & 7.99 & 7.57 & 7.24 & 6.94 & 6.68 & 14.99 & 84.94 \\
    Variant 2  &  79.09  & 74.27 & 69.83 & 65.77 & 62.53 & 59.79 & 57.22 & 55.08 & 52.63 & 50.65 & 48.82 & 61.42 & 7.63 \\
    Variant 3 & 79.29 & 13.64 & 10.45 & 9.80 & 9.41 & 8.65 & 8.42 & 7.90 & 7.28 & 7.26 & 6.83 & 15.36 & 85.27  \\
    Variant 4 & 79.29 & 73.64 & 70.31 & 67.07 & 65.04 & 63.15 & 60.78 & 59.03 & 56.90 & 55.58 & 54.10 & 64.08 & 6.92  \\
    \midrule
    IR$_r$ & 79.29 & 75.54 & 72.22 & 69.19 & 66.98 & 65.16 & 62.71 & 61.05 & 58.84 & 57.57 & 56.11 & 65.88 & 6.89 \\
    \bottomrule
    \end{NiceTabular}
    }
\end{table*}

\subsection{Overall Result}
Following the advocacy of the CIL community~\citep{zhou2023deep}, we compare the memory budgets of all comparisons. ResNet-18 consumes $11,359,769$ floating-point parameters for both CIFAR100 and ImageNetSubset (100 classes), and $11,462,369$ parameters for TinyImageNet (200 classes). The memory footprint for each image is in integer values. Therefore, exemplar-based methods (iCaRL-CNN, iCaRL-CNN, WA) demand approximately $11,359,769$ floats $\times 4$ bytes/float $\approx 43.33$ megabytes (MB) for model storage and an additional $(3\times32\times32)$ bytes/image $\times2000$ images $\approx 5.86$ MB for CIFAR100 exemplars. Note that DER incures a memory budget of $(P+1)\times43.33+5.86$ MB, where $P$ is the number of incremental phases. In contrast, efficient CIL methods (EWC, lwf.MC, PASS, IL2A, SSRE, FetrIL, IR) utilize a more frugal $43.33$ MB exclusively for model storage, showcasing efficiency, particularly on TinyImageNet and ImageNetSubset datasets.

\textbf{Absolute Accuracy: IA$\uparrow$.}\quad
Table~\ref{tab:ia} shows the proposed incremental representation framework achieves comparable results with other efCIL methods, demonstrating the effectiveness of the IR framework without storing old exemplars or constructing old features. For instance, in $10$ phases, IR$_r$ performs similarly to the most recent efCIL method, FetrIL~\citep{petit2023fetril}, achieving $65.88\%$ versus $64.44\%$ on CIAFR100, and $52.55\%$ versus $50.77\%$ on TinyImagenet. Additionally, IR outperforms the strong baseline iCaRL-NME~\citep{rebuffi2017icarl} by $8.27\%$ on CIFAR100 ($20$ phases) and performs comparably to state-of-the-art exemplar-based methods which use many saved exemplars overall. Similar observations hold for the ImageNetSubset~\footnote{We have observed a performance disparity between our reproduced FetrIL and that presented in the original paper~\citep{petit2023fetril}. This primarily attributes to the inclusion of additional operations, such as cutout and colorJitter~\citep{zhou2023deep}.}.

\textbf{Relative Forgetting: F$\downarrow$.}\quad
As depicted in Figs.\ref{fig:fgt_cifar}-\ref{fig:fgt_imagenet100_10}, our results show that incremental representation methods, IR$_r$ and IR$_m$, notably mitigate the forgetting problem compared with established exemplar-based methods. It confirms the effectiveness of the IR framework in facilitating model memorization, even in the absence of storing old exemplars. For instance, considering the final forgetting measure across $10$ incremental phases, our methods outperform the best exemplar-based approach DER by gaps of $15.97\%$ on CIFAR100, $12.29\%$ on TinyImageNet, and $6.73\%$ on ImageNetSubset. Furthermore, our methods surpass the state-of-the-art efCIL methods by a large margin, further confirming the effectiveness of the IR framework without requiring the construction of old features.

In conclusion, the IR framework is good enough to achieve CIL without storing exemplars or constructing old pseudo-features. It is comparable in achieving incremental accuracy and remarkably superior in preventing the model from forgetting, compared with the exemplar-based CIL methods and existing efCIL methods that rely on elaborately constructing old pseudo-features.

\subsection{Ablation Study}
IR comprises three components: AUG, transient CE (Tra.CE), and space maintenance (Spa.M), as shown in Fig.\ref{fig:method}. To evaluate the effect of each component, we conduct the ablation study and present the results for both $5$ and $10$ phase settings on CIFAR100 in Tab.\ref{tab:abl_5} and Tab.\ref{tab:abl_10}, respectively. 
We observe: 
(1) Tra.CE alone fails in class incremental learning without Spa.M and AUG. (2) AUG's effect in combination with Tra.CE is relatively small due to the severe drifting of learned representations from previous tasks. (3) Spa.M successfully mitigates the drifting problem and outperforms Tra.CE. For instance, Variant2 improves Variant1's IA by 39.80\% on $5$ phases and 44.43\% on $10$ phases. (4) Combining Spa.M with AUG further enhances performance, particularly in long phases. For example, Variant4 improves Variant2 by $2.66\%$ in IA and $0.74\%$ in Average F (Avg.F) on $10$ phases, and by $0.77\%$ in IA and $0.71\%$ in Avg.F on $5$ phases. Moreover, IR$_r$ maintains model stability against forgetting while improving Variant4's IA performance, demonstrating the effectiveness of Tra.CE and Spa.M with AUG. It also indicates that Tra.CE and Spa.M can mutually benefit from each other under the AUG augmentation.

\subsection{Method Analysis}
For our method analysis, we conduct experiments on CIAFR100 with $4$ tasks with an equal number of classes. 

\begin{figure}[!t]
    \centering
    \subfigure[IR$_r$]{\label{fig:para_r}\includegraphics[width=0.49\linewidth]{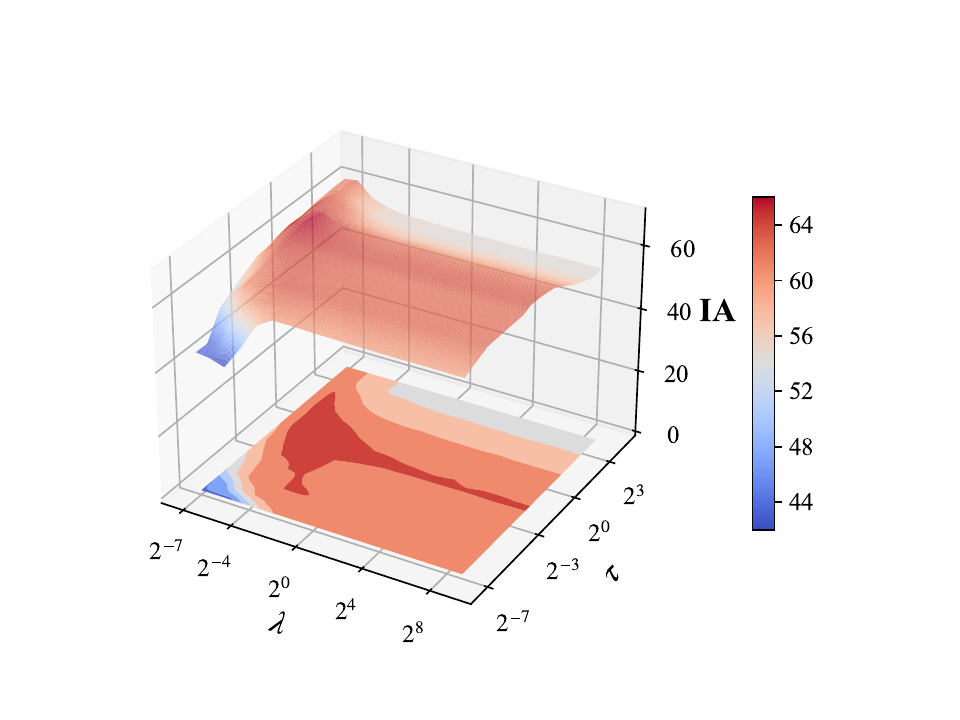}}
    \hspace*{\fill}
    \subfigure[IR$_m$]{\label{fig:para_m}\includegraphics[width=0.47\linewidth]{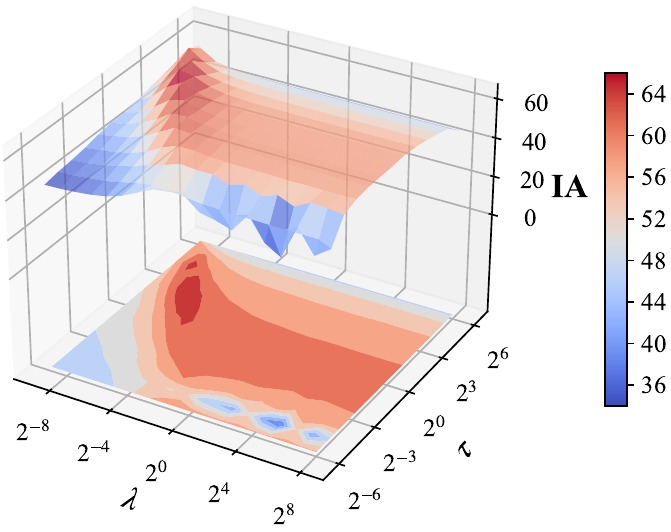}}
    \caption{Sensitivity results of hyper-parameters, temperatures $\tau$ and loss trade-off $\lambda$, in (a) IR$_r$ and (b) IR$_m$. IR robustly achieves competitive performance, and relatively larger $\tau$ and smaller $\lambda$ are superior (\textit{best view in color}). }
    \label{fig:para}
\end{figure}

\textbf{Parameter Tuning.}\quad
Two parameters are needed to tune in IR, the temperature $\tau$ and the trade-off parameter $\lambda$. We conduct $\tau$ varies from $2^{-7}$ to $2^8$ while $\lambda$ changes from $2^{-7}$ to $2^3$ for IR$_r$, and $\tau\in\{2^{-9}\sim2^8\}$ while $\lambda\in\{2^{-6}\sim2^6\}$ for IR$_m$. 
Fig.\ref{fig:para} shows that IR achieves competitive performance (with IA over $60\%$) robustly under a wide range of hyper-parameter values, i.e., $2^{-3}\leq\tau\leq2^4$ and $2^{-4}\leq\lambda\leq2^8$, with either rotation or mixup as AUG strategy, which verify the stability of our method. Besides, we observe that, relatively flat and plastic representations (controlled by larger temperature and smaller trade-off parameters, respectively, as discussed in Section \ref{subsec:proposed}) are superior for CIL.

%

\begin{figure}[!t]
\centering
\includegraphics[width=0.98\linewidth]{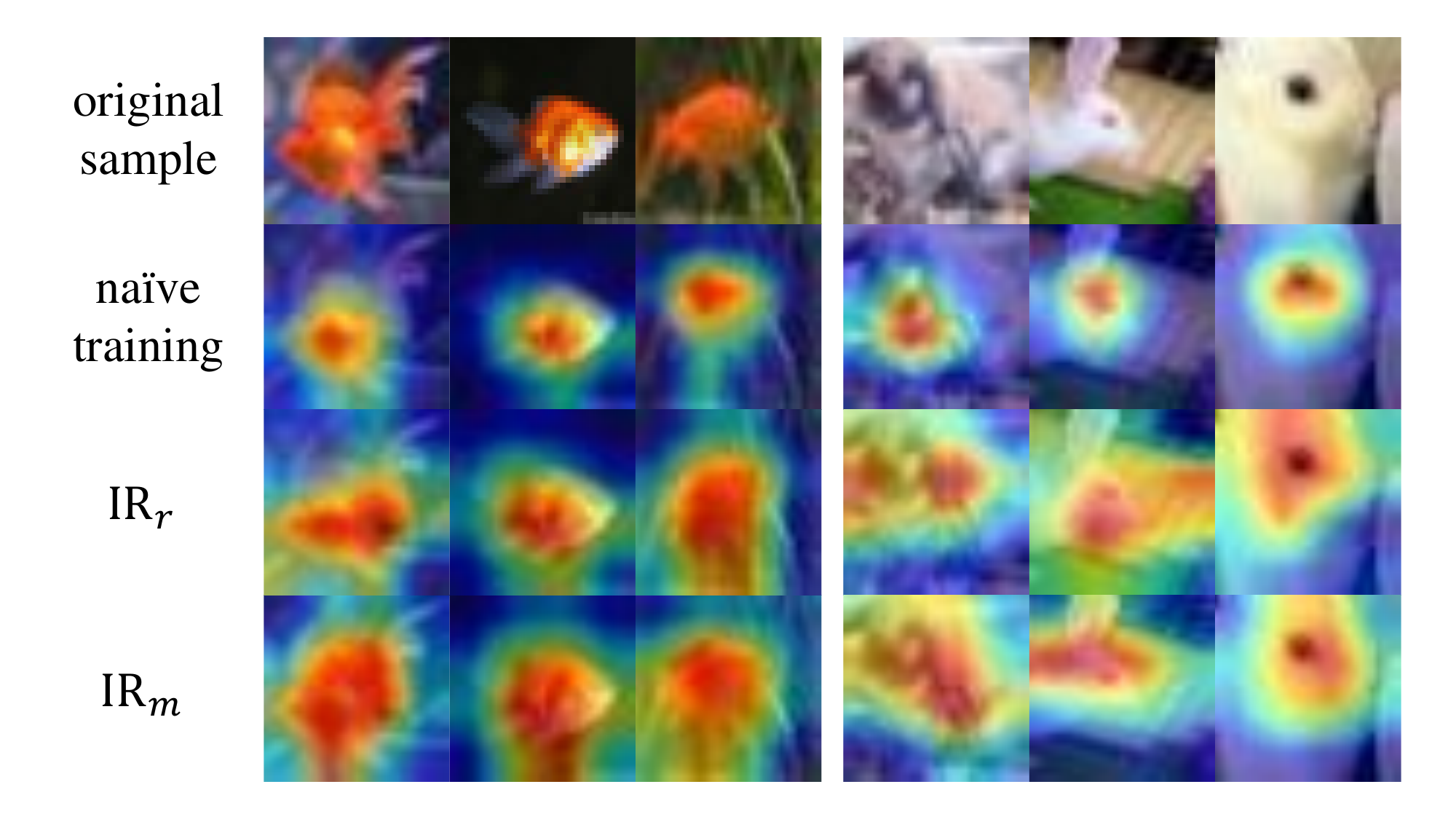}
\caption{Feature space indicated with the gradient-weighted class activation mapping of various training methods. IR with rotation or mixup AUG enables the network to gain more attention area than na\"ive training.}
\label{fig:cam}
\end{figure}

\textbf{Space Enlargement Achievement.}\quad
We engage the gradient-weighted class activation mapping~\citep{selvaraju2020grad} to highlight the network-focused localization on the original images. We randomly select $2$ of $25$ initial classes and visualize them using the well-trained model with naive training and our IR$_r$ and IR$_m$. As shown in Fig.\ref{fig:cam}, naive training
only focuses on a few part vision locations of the fish or rabbit. In contrast, IR with either rotation or mixup strategies expands the network's attention area in different forms encompassing the focused region of na\"ive training. Overall, it verifies that rotation or mixup enlarges the feature space by expanding the network's attention area. 

\begin{table}[!t]
  \caption{Space maintenance results using different strategies, i.e., L1-, L2-, and Nuclear-norm. We report the average accuracy of the initial phase (ini.$0$) and the $10$ incremental phases (inc.$x$), the average incremental accuracy (IA$\uparrow$), and the average forgetting measure (Avg.F$\downarrow$).}
  \label{tab:Lp}
  \centering
  \resizebox{.998\linewidth}{!}{
    \begin{NiceTabular}{cc | llll | ll}[code-before = \rowcolors{}{}{Gray} \rowcolor{TitleColor}{1}]
    \toprule
    \multicolumn{2}{c}{Strategy} & ini.$0$ & inc.$1$ & inc.$2$ & inc.$3$ & IA$\uparrow$ & Avg.F$\downarrow$ \\
    \midrule
    \multirow{3}{*}{ IR$_r$ } 
    & L1 & 88.55 & 65.17 & 55.63 & 48.94 & 64.57 & 13.83 \\
    & L2 & 88.55 & 65.18 & 55.54 & 48.95 & 64.55 & 13.94 \\
    & Nuclear & 88.55 & 65.15 & 55.55 & 48.75 & 64.50 & 13.98 \\
    \midrule
    \multirow{3}{*}{ IR$_m$ } 
    & L1 & 88.25 & 64.75 & 54.13 & 46.19 & 63.83 & 14.75 \\
    & L2 & 88.25 & 64.87 & 54.22 & 46.37 & 63.93 & 14.85 \\
    & Nuclear & 88.25 & 64.71 & 54.05 & 46.22 & 63.81 & 14.91 \\    
    \bottomrule
    \end{NiceTabular}
    }
\end{table}
\textbf{Maintenance Strategy.}\quad
The space maintenance loss of Eq.\eqref{eq:sm} is realized with the $L_p$-norm regularization. We engage three commonly used norms, $L_1$-, $L_2$-, and $Nuclear$-norm to IR$_r$ and IR$_m$. As shown in Tab.\ref{tab:Lp}, IR$_r$ and IR$_m$ are robust to different maintenance strategies. For example, IR$_r$ achieves similar IA$\uparrow$ of L1, L2, and Nuclear with $64.57\%$, $64.55\%$, and $64.50\%$ (the corresponding Avg.F$\downarrow$ are $13.83\%$, $13.94\%$, and $13.98\%$), respectively. 

\section{Conclusion}
This paper proposed an exemplar-free class incremental learning (efCIL) framework called incremental representation (IR). Unlike recently popular efCIL methods that require constructing pseudo-features and incrementally training the top classifier, IR simplifies the process by maintaining a proper feature space with neither pseudo features nor incrementally training classifiers. The feature space construction is achieved by dataset augmentation strategies, while maintenance is ensured through the simple L2 loss. We demonstrated the effectiveness of IR on various benchmark datasets using rotation (IR$_r$) and mixup (IR$_m$) strategies. Our results showed that IR achieves comparable incremental accuracy and superior forgetting measures with the state-of-the-art efCIL methods. Considering that the mainstream of related work is generally based on generating elaborate pseudo features, we hope our work can shed some light on the efCIL community.

\nocite{langley00}
\newpage
Impact Statements: This paper presents work whose goal is to advance the field of Machine Learning. There are many potential societal consequences of our work, none which we feel must be specifically highlighted here.
\bibliography{example_paper}
\bibliographystyle{icml2024}

\newpage



\end{document}